\pdfoutput=1

\documentclass[11pt]{article}

\usepackage[preprint]{acl}

\usepackage{times}
\usepackage{latexsym}

\usepackage[T1]{fontenc}

\usepackage[utf8]{inputenc}

\usepackage{microtype}

\usepackage{inconsolata}

\usepackage{graphicx}

\usepackage{booktabs} 
\usepackage{multirow} 
\usepackage{pdflscape} 
\usepackage{graphicx}
\usepackage{amsmath} 
\usepackage{tabularx}
\usepackage{subcaption} 
\usepackage{amsmath}

\usepackage{wrapfig}
\usepackage{seqsplit}
\usepackage{marvosym} 
\usepackage{textcomp}

\title{DAST: Difficulty-Adaptive Slow Thinking for Large Reasoning Models }

\author{
Yi Shen$^{1,2}$, Jian Zhang$^{1,2}$, Jieyun Huang$^{1,2}$,  \Letter Shuming Shi$^{1,2}$, Wenjing Zhang$^{1,2}$, \\
\textbf{Jiangze Yan$^{1,2}$,} \textbf{Ning Wang$^{1,2}$,} \textbf{Kai Wang$^{1,2}$,} \textbf{Zhaoxiang Liu$^{1,2}$,}  \Letter \textbf{Shiguo Lian}$^{1,2}$ \\
$^{1}$Unicom Data Intelligence, China Unicom \quad \\
$^{2}$Data Science \& Artificial Intelligence Research Institute,  China Unicom \quad \\
\Letter \ Corresponding Authors\quad \\
\texttt{\{sheny73, zhangj2791,liansg\}@chinaunicom.cn, ssm01@hotmail.com}\\
}

\begin{document}

\maketitle

\begin{abstract}

Recent advancements in slow thinking reasoning models have shown exceptional performance in complex reasoning tasks. However, these models often exhibit overthinking (generating redundant reasoning steps for simple problems), leading to excessive computational resource usage. While current mitigation strategies uniformly reduce reasoning tokens, they risk degrading performance on challenging tasks that require extended reasoning. This paper introduces Difficulty-Adaptive Slow Thinking (DAST), a novel framework that enables models to autonomously adjust the length of Chain-of-Thought (CoT) based on problem difficulty. We first propose a Token Length Budget (TLB) metric to quantify difficulty, then leverage budget-aware reward shaping and budget preference optimization to implement DAST. DAST penalizes overlong responses for simple tasks while incentivizing sufficient reasoning for complex problems. Experiments on diverse datasets and model scales demonstrate that DAST effectively mitigates overthinking (reducing token usage by over 30\% on average) while preserving reasoning accuracy on complex problems. Our codes and models are available at \url {https://github.com/AnonymousUser0520/AnonymousRepo01}.

\end{abstract}

\section{Introduction}

Recently, significant advancements have been made in slow-thinking reasoning models, exemplified by OpenAI's o1 \cite{o1} and DeepSeek-R1 \cite{guo2025deepseekr1}. These slow-thinking reasoning models, which simulate human deep-thinking mechanisms through self-reflection, error correction, and exploration of multiple solution strategies, have demonstrated remarkable potential in complex reasoning tasks such as  mathematical problem-solving \cite{AIME2024} and programming challenges \cite{jain2024livecodebench}.

\begin{figure}
\centering
\includegraphics[width=0.5\textwidth]{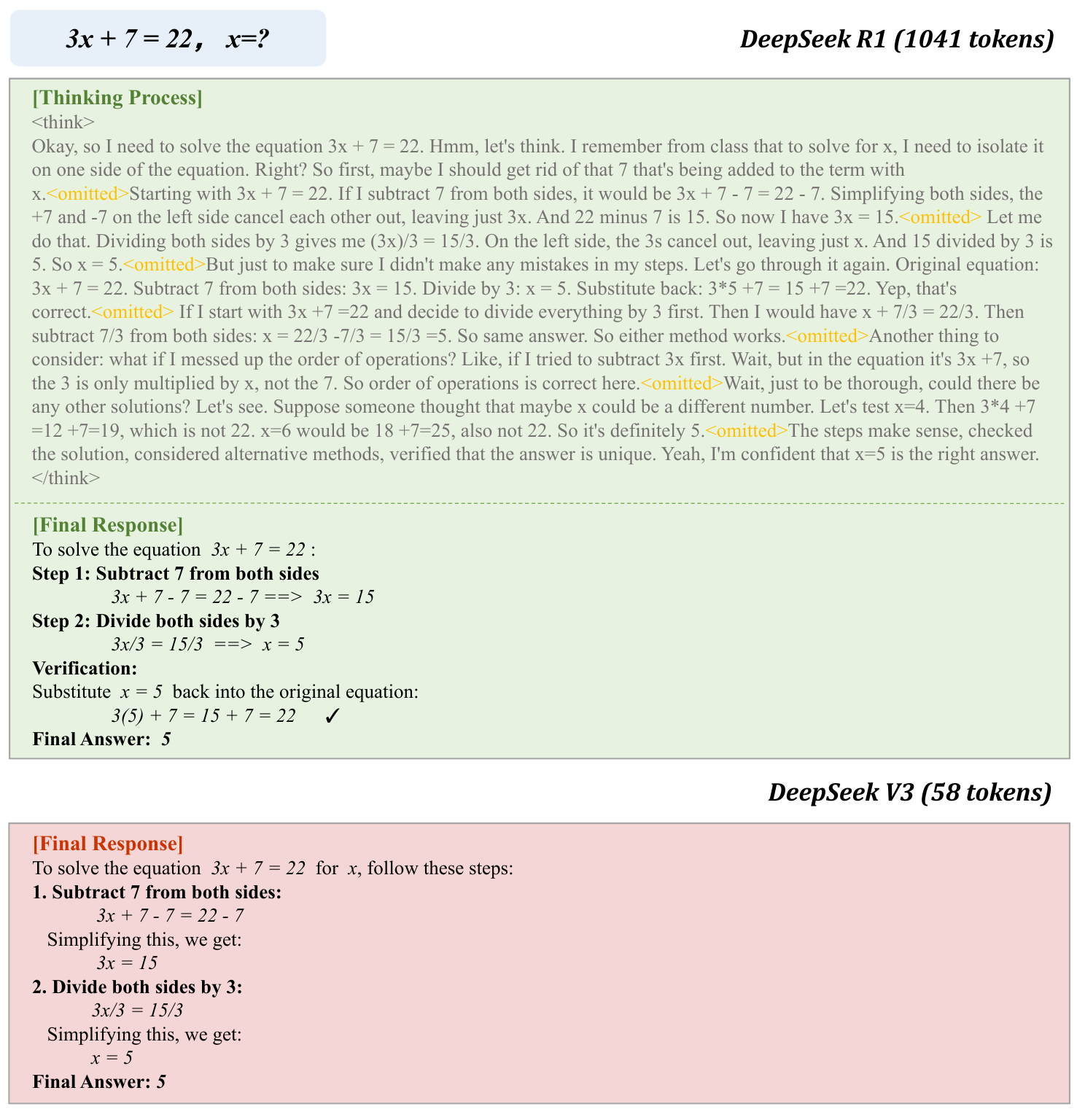}
\caption{An Example to illustrate the overthinking phenomenon.} \label{fig:overthink}
\end{figure}

However, empirical studies \cite{chen20242add3,2025Stop,Liu2025EfficientIF} have shown that these reasoning models suffer from overthinking phenomenon. In other words, these models tend to generate redundant solutions and unnecessarily complex reasoning steps when addressing simple problems, leading to inefficient computational resource utilization. For instance, as demonstrated in Figure \ref{fig:overthink}, traditional LLM (DeepSeek V3) can solve basic mathematic problems such as ``\textit{3x + 7=22, x=?}'' with just 58 tokens, while reasoning model such as DeepSeek-R1  may consume over 1000 tokens for the same problem. This overthinking phenomenon not only significantly reduces reasoning efficiency but also causes information overload for the users.

Current approaches \cite{2025Stop,xia2025tokenskip,chen20242add3} to mitigate the overthinking problem typically employ a one-size-fits-all strategy, uniformly reducing reasoning steps or token counts across all problems. Although these approaches significantly reduce the output length of slow-thinking models, they carry the risk of performance degradation, particularly when addressing challenging problems. Prior studies \cite{zeng2024scaling,muennighoff2025s1} have demonstrated that adequate reasoning length is critical for slow-thinking models to effectively solve complex tasks. It is therefore essential to devise approaches that mitigate overthinking phenomena while maximally preserving reasoning capabilities.

This raises a fundamental question: Can slow thinking models autonomously adjust reasoning depth based on problem difficulty, thereby generating concise responses for simple questions while maintaining sufficiently extended CoT reasoning for complex ones? We propose a Difficulty-Adaptive Slow-Thinking (DAST) framework to tackle this challenge.

Our key idea is straightforward: Given that tasks of varying difficulty levels inherently demand different reasoning depths, we propose to establish a mapping relationship between problem complexity and target response length. By comparing the length of the current response with the target response length, we can determine whether to apply additional rewards or penalties to the current answer. Building upon this, we construct a training objective to achieve adaptive reasoning. Specifically, we first introduce a difficulty quantification metric termed ``Token Length Budget'' (TLB), which integrates both the accuracy of sampled responses and their length distributions. This metric effectively combines problem difficulty characteristics with token length information. For multiple generated responses sampled, our method applies budget-aware reward shaping: Responses exceeding the TLB of simple questions receive penalty signals, while those approaching the TLB for complex problems receive positive incentives. This mechanism allows us to construct pair-wise budget  preference training data that inherently encodes the relationship between problem difficulty and target response length. Through follow-up preference optimization, we enable the slow-thinking model to acquire adaptive reasoning capabilities, strategically allocating more computational resources to challenging problems while maintaining efficient processing of simpler tasks. The proposed DAST method essentially establishes a learnable mapping between problem difficulty levels and corresponding target response length, achieving intelligent computation allocation during the inference stage without compromising reasoning quality.

Our main contributions are as follows:
\begin{enumerate}
    \item We propose a difficulty-adaptive slow thinking (DAST) scheme, which effectively alleviates the phenomenon of overthinking while maintaining the reasoning performance, especially on difficult tasks.    
    \item We propose a novel problem difficulty quantification metric (TLB) that is applicable to many downstream tasks.
    \item We conduct extensive validation experiments across multiple datasets with models of varying parameter scales. The results demonstrate that the proposed DAST approach effectively mitigates the overthinking problem while preserving the model's reasoning capabilities.
\end{enumerate}

\section{Related Work}


\subsection{Problem Difficulty Evaluation }
Some previous studies mainly used proprietary models like ChatGPT to assess the difficulty or complexity of data samples \cite{lu2023instag,liu2023makes}. Such methods are limited by the ability of the LLMs they employed for evaluation. In reasoning scenarios such as mathematics, a more common solution is to use sampling accuracy to measure the difficulty of the problem \cite{tong2025dart,team2025kimi}. However, this approach has two shortcomings. First, it requires sampling more answers to ensure effectiveness. Second, for some extremely difficult questions, there are fewer valid answers, resulting in insufficient discrimination. The token length budget metric we propose in this paper can effectively circumvent these two shortcomings.

\subsection{Overthinking }


Mitigating ``overthinking'' in Large Reasoning Models (LRMs) to enhance reasoning efficiency has garnered increasing research attention. Existing approaches can be broadly categorized into three main types \citep{2025Stop,Liu2025EfficientIF}: prompt-based methods, output-based methods, and post-training methods.

\noindent\textbf{Prompt-based Methods.} These methods focus on modifying input prompts to achieve concise reasoning. Common techniques include imposing explicit token limits \citep{Nayab2024ConciseTI} or instructing the model to generate more succinct reasoning chains \citep{xu2025chainofdraft,renze2024ccot}.


\noindent\textbf{Output-based Methods.} These approaches intervene during the inference stage to control the reasoning process. One line of work compresses intermediate reasoning steps into latent representations rather than explicit text, thereby enhancing brevity at the cost of interpretability \citep{hao2024coconut, shen2025codi}. Another direction involves dynamic decoding, wherein lightweight models or heuristics evaluate each reasoning step during generation, deciding whether to retain, modify, or discard it \citep{sun2024speculativerejection, Yang2025DynamicEE,zhang2025lightthinker}.

\noindent\textbf{Post-training Methods.} These methods typically involve supervised fine-tuning (SFT) on variable-length Chain-of-Thought (CoT) data \citep{chen20242add3,ma2025cotvalue,kang2025c3ot,xia2025tokenskip,munkhbat2025self} or the incorporation of length-based rewards within reinforcement learning frameworks \citep{team2025kimi,luo2025o1,arora2025training,yeo2025demystifying}. The objective is to guide the model towards generating reasoning paths that are both concise and accurate.

While all these methods have demonstrated promising results in efficient reasoning, they exhibit certain limitations. Most existing approaches indiscriminately compress the Chain-of-Thought (CoT) across all problems, leading to degraded reasoning performance, particularly for complex problems. Furthermore, the majority of these methods have primarily been evaluated on LRMs with fewer than 7B parameters, with limited exploration of their efficacy on models with larger parameter scales (e.g., 32B). Our work aims to achieve difficulty-adaptive efficient inference by defining a token length budget for each problem, reflecting its perceived difficulty for the current model. Although some prior studies \cite{aggarwal2025l1,muennighoff2025s1} have employed token budgets to control inference length, these approaches typically rely on pre-defined, discrete token budgets, often manually set. The most closely related work is that of \cite{han2024tokenbudget}, which also attempts to allocate varying token budgets based on problem complexity. However, their method necessitates an iterative search process, experimenting with token limits within prompts to determine the final budget, which can be cumbersome. Additionally, its effectiveness has not been demonstrated on ``slow thinking'' models.

\begin{figure*}[t]
\centering
\includegraphics[width=0.95\textwidth]{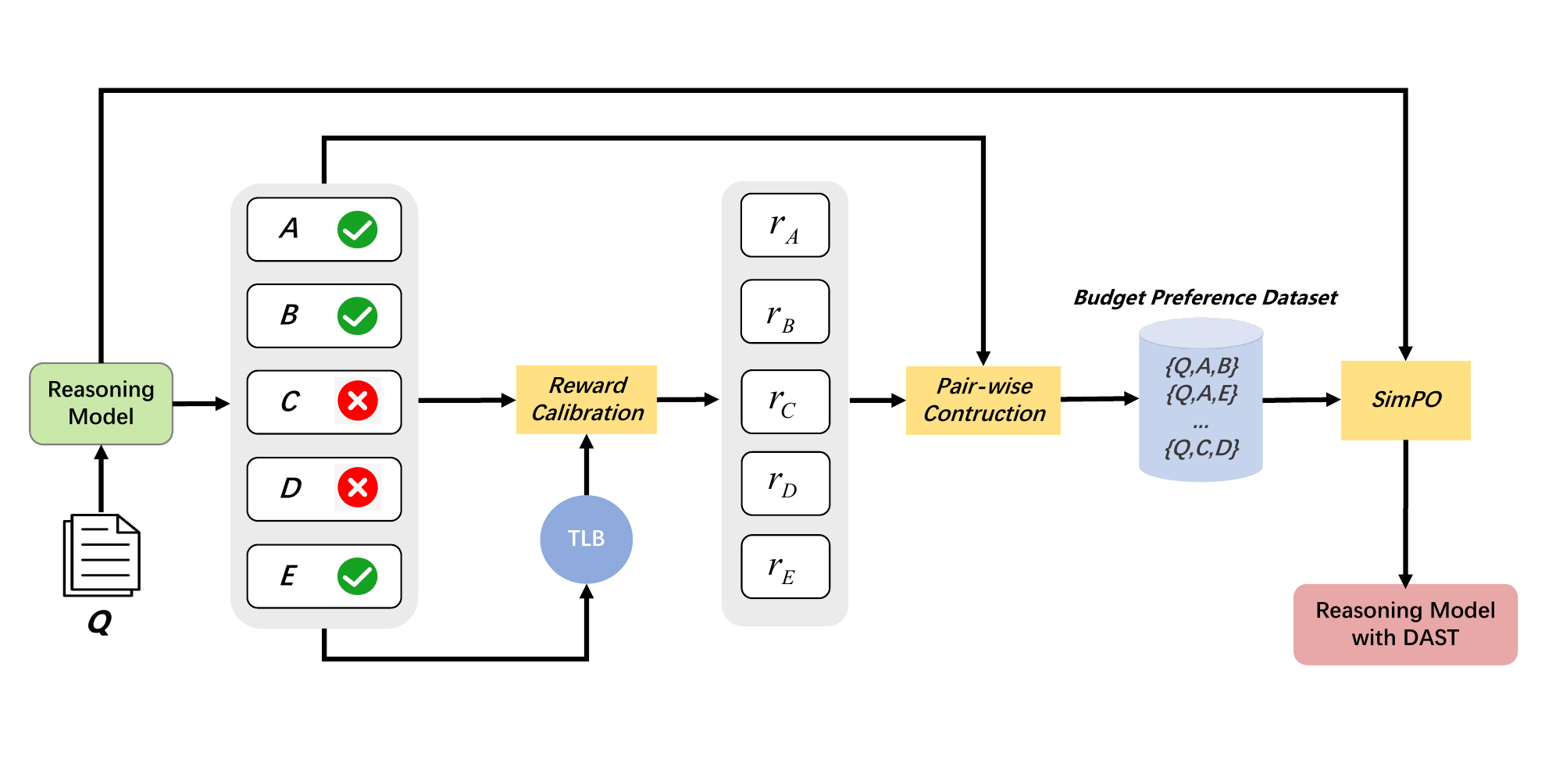}
\caption{Overview of our proposed DAST method.} \label{fig:dast}
\end{figure*}

\section{Methodology}

In this section, we introduce our proposed DAST method in detail. Our key insight lies in enhancing existing reasoning models through budget-preference training, enabling adaptive response generation with length that corresponds to problem complexity. The main challenge lies in establishing a principled relationship between response length and problem difficulty. To this end, we propose a novel reasoning Token Length Budget (TLB) metric that dynamically scales with problem complexity: simpler questions receive smaller length allocations while complex ones are allocated extended budgets. This metric not only serves as a length reference for response generation but also could be used to quantify problem difficulty.

The technical implementation of DAST involves three crucial steps: First, calibrating the initial rule-based reward scores of each response with thinking process via comparing its actual token length with the TLB of the corresponding problem. Second, constructing a pairwise budget-preference training dataset based on the calibrated reward scores. Finally, employing SimPO \cite{meng2025simpo} to fine-tune the original reasoning model, endowing it with adaptive reasoning capabilities. The overall framework of DAST is depicted in Figure \ref{fig:dast}.

\subsection{Token Length Budget Definition}

Our proposed  Token Length Budget (TLB) metric is formally defined as:

{\small
\begin{equation}
L_{budget} = p\cdot L_{\overline{r}} + ( 1-p )\cdot L_{max} ,
\label{equa:TBL}
\end{equation}
}
where {\small $$
p = \frac{c}{N}
$$} denotes the sampling accuracy. Here, $c$ is the number of correct responses sampled from the current question with the backbone LRM, $N$ is the total number of sampled responses. $L_{\overline{r}}$ represents the average token length of correct responses, and $L_{\mathrm{max}}$ is the maximum generation length.

\begin{figure}
\centering
\includegraphics[width=0.5\textwidth]{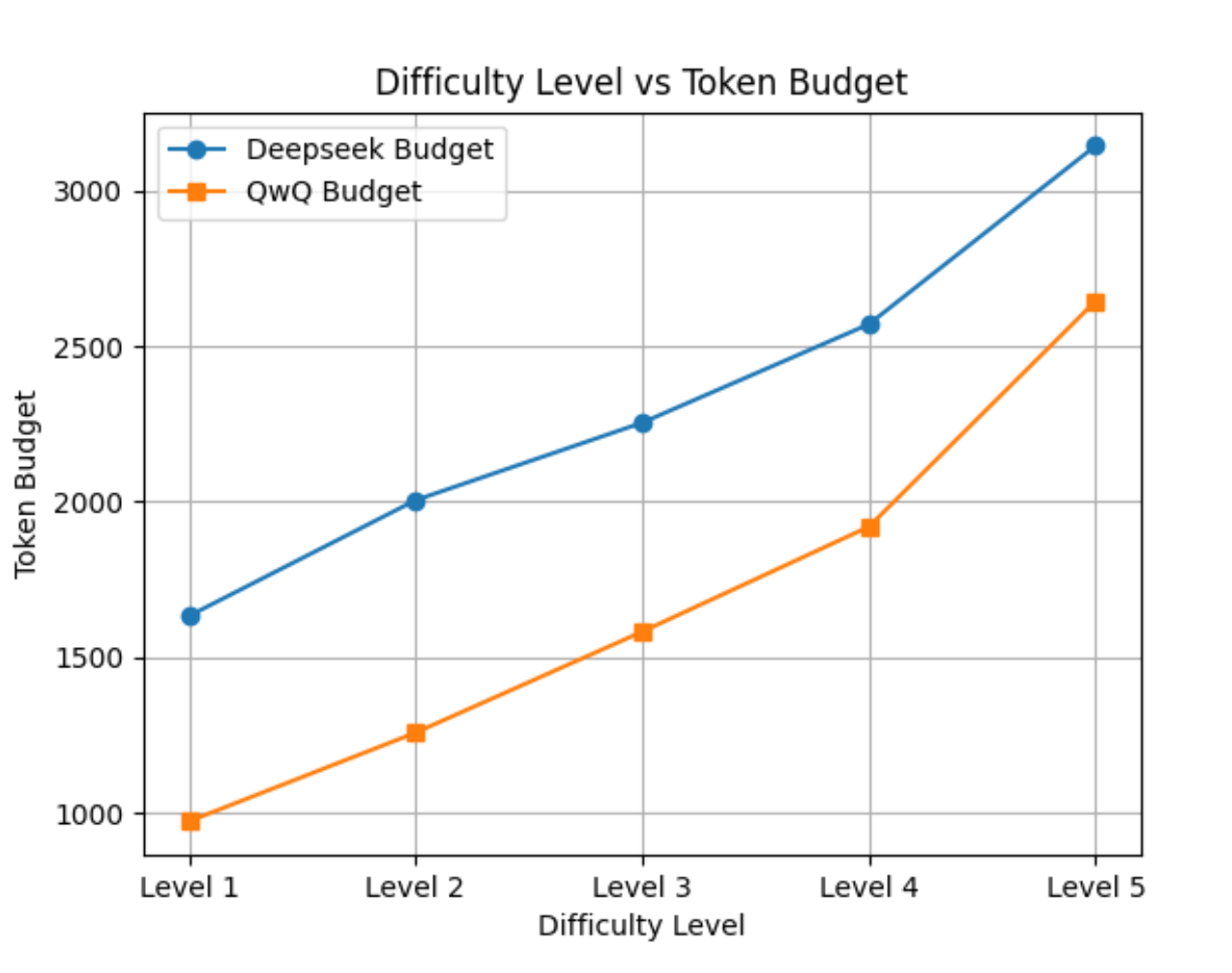}
\caption{Average token budget distribution across difficulty levels ($L_{\mathrm{max}} = 4096$). Results are computed using responses sampled from QwQ-32B-preview and DeepSeek-R1-Distill-Qwen-32B (DS-32B) on the MATH training set. The higher TLB for DS-32B stems from its structured output format containing both reasoning chains and final answers. } \label{fig2:tlb}
\end{figure}

The higher the sampling accuracy, the closer $L_{budget}$ is to the average length of correct responses ($L_{\overline{r}}$), while lower accuracy brings $L_{budget}$ closer to the maximum generated length. A sampling accuracy of 0 indicates extreme difficulty, in which case the model should be encouraged to think deeply and generate longer CoT. At this point, TLB equals the model’s maximum generation length. As shown in Figure \ref{fig2:tlb}, the average TLB exhibits strong positive correlation with problem difficulty level on the MATH training dataset, demonstrating its potential as an effective measure for quantifying problem complexity.

\subsection{Reward Score Calibration}
\label{subsection:reward}

In reasoning scenarios such as mathematics and coding, o1-like slow thinking models typically employ rule-based rewards as feedback signals for training \cite{guo2025deepseekr1,team2025kimi}. In this work, traditional rule-based rewards are calibrated by incorporating the deviation between actual response length and the TLB metric. This calibration allows the reward score to jointly capture both difficulty-aware information and length characteristics, enabling difficulty-adaptive training.

\begin{equation}
\resizebox{0.95\columnwidth}{!}{%
$\mathit{reward}(i) = 
\begin{cases}
    \max(-0.5\lambda + 0.5, 0.1) & \text{if correct} \\
    \min(0.9\lambda - 0.1, -0.1) & \text{if incorrect},
\end{cases}$%
}
\label{equa:reward}
\end{equation}

where 
\[
\lambda = \dfrac{L_{i}-L_{budget}}{L_{budget}}
\]

\begin{figure}[htbp]
\centering
\includegraphics[width=0.5\textwidth]{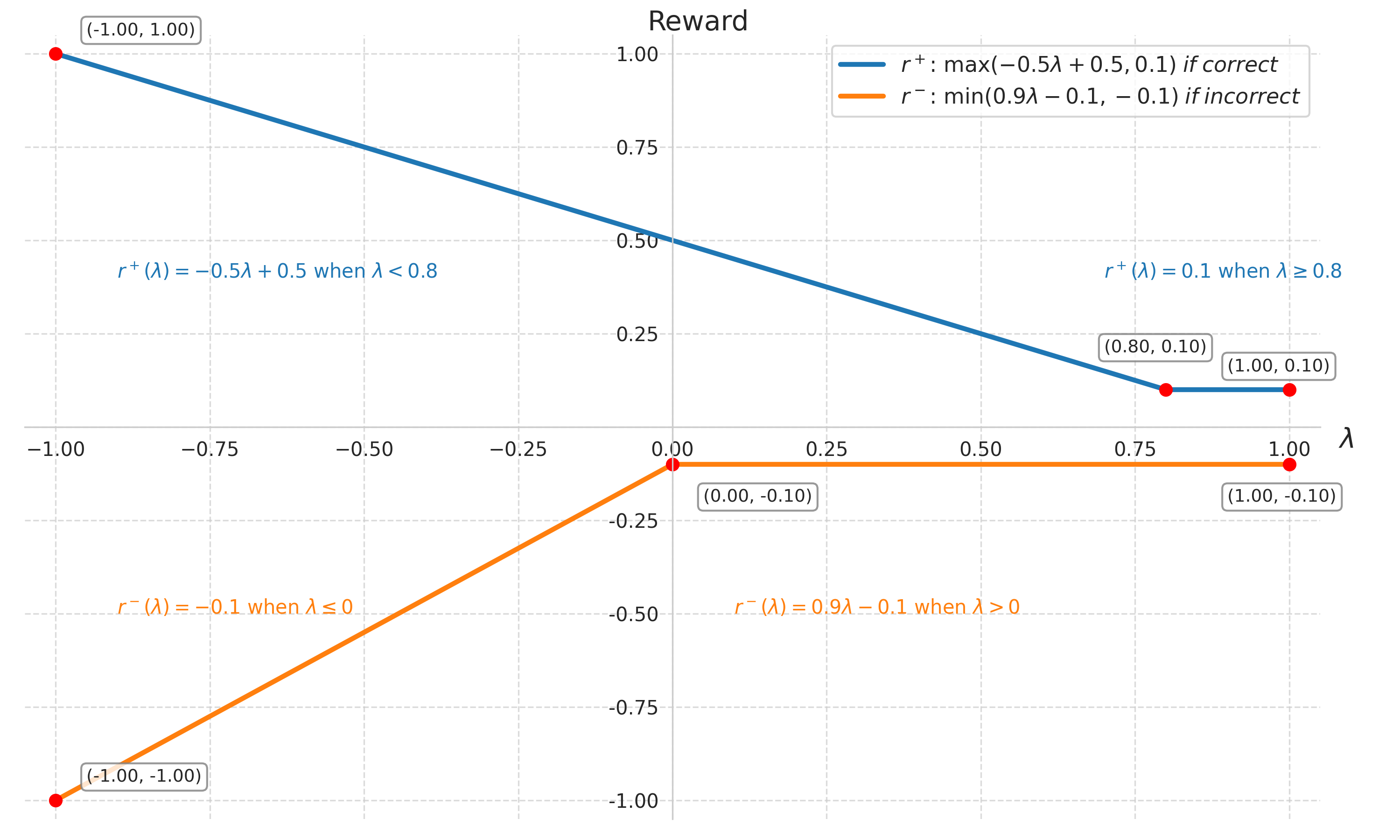}
\caption{Calibrated reward function with TLB.} \label{fig:reward}
\end{figure}

The calibrated reward score for the response $i$ is defined as Equation 2. From Figure \ref{fig:reward}, we can derive the following insights:

For a correct answer, if its length exceeds TLB, it will result in a reward decay. The simpler the question, the smaller the TLB. If the generated length significantly surpasses TLB, the reward will decay severely. Conversely, if it falls below TLB, the reward will be amplified, encouraging the model to generate shorter answers within the corresponding TLB.

For incorrect answers, if the actual length is below TLB, it indicates insufficient reasoning. In this case, the model is encouraged to engage in more thorough thinking process and generate longer responses. The closer the length is to TLB, the greater the calibrated reward score. Once the TLB is reached, the reward score saturates.

\subsection{Budget Preference Data Construction}

For each input question $x$, $N$ candidate responses are sampled with corresponding TLB $L_{\mathrm{budget}}^{(x)}$ computed as formalized in  Equation \ref{equa:TBL}. The corresponding reward scores are then derived using Equation \ref{equa:reward}.  These responses are subsequently ranked based on their reward scores to constructed contrastive pairs $(x, y_w, y_l)$ for subsequent preference optimization, where $y_w$ and $y_l$ denote the winning and losing responses respectively.

We categorize contrastive pairs into two distinct classes \footnote{There is actually a third class: Correct-InCorrect Pair (CICP), but our experiments show that CICP does not improve the  performance. We will discuss the effects of CICP in Section \ref{subsection:ablation} }: 
(1) Dual-Correct Pair (DCP): Both responses yield correct answers, but the preferred instance $y_w$ demonstrates significantly higher output conciseness ($|y_w| \ll |y_l|$). DCP is designed to encourage the model to generate responses that are both correct and as concise as possible within the token length budget. 
(2) Dual-InCorrect Pair (DICP): Both responses produce incorrect answers, yet $y_w$ exhibits substantially longer reasoning chains ($|y_w| \gg |y_l|$). DICP is designed to stimulate more extensive reasoning attempts when the model has not yet produced a correct answer and  remains within the corresponding TLB.

For each question, we first select the DCP and DICP pairs with maximal reward margin $\Delta R = R(y_w) - R(y_l)$, then apply a two-stage filtering process:  
1. We establish a truncation threshold $\delta \in (0,1)$ to eliminate the bottom $\delta|D|$ pairs with minimal $\Delta R$, where $|D|$ denotes the candidate set size.  
2. To maintain data quality and training efficiency, we retain at most two highest-margin pairs (one DCP and one DICP) per question.

This selection mechanism ensures statistical significance in reward differences while preserving informative contrastive signals, ultimately enhancing the stability of preference optimization.

\subsection{Budget Preference Training}

The constructed dataset $\mathcal{D}_{\mathrm{pre}}$ enables alignment of reasoning LLMs through Simple Preference Optimization (SimPO) \cite{meng2025simpo}. We chose SimPO due to its characteristic of being more sensitive in controlling answer length. The optimization objective is formulated as:

{
\small
\begin{multline}
\mathcal{L}_{\mathrm{SimPO}}(\pi_\theta) = -E_{(x, y_w, y_l) \sim \mathcal{D}} \Bigl[ \log \sigma \Bigl( \\
\frac{\beta}{|y_w|} \log \pi_\theta(y_w | x) - \frac{\beta}{|y_l|} \log \pi_\theta(y_l | x) - \gamma \Bigr) \Bigr],
\end{multline}
}

where $\beta$ and $\gamma$ are hyperparameters.

\section{Experiments}

\subsection{Experimental Setup}

\paragraph{Backbone Reasoning Models}
We conduct comparative experiments on two Large
Reasoning Models (LRMs): DeepSeek-R1-DistillQwen-7B (DS-7B) and DeepSeek-R1-Distill-Qwen-32B (DS-32B) \cite{guo2025deepseekr1}. Although both models exhibit
strong reasoning abilities, their substantial redundant thought processes highlight the necessity of response compression.

\paragraph{Benchmarks}


We evaluate model performance on three established reasoning benchmarks:

\begin{itemize}
    \item \textbf{MATH-500} \cite{lightman2023let}: 500 high school competition-level mathematical problems stratified into 5 difficulty levels, ranging from Level 1 (easiest) to Level 5 (hardest);
    \item \textbf{AIME 2024} \cite{AIME2024}: 30 curated problems from the American Invitational Mathematics Examination testing complex problem-solving;
    \item \textbf{GPQA} \cite{rein2024gpqa}: 198 PhD-level science questions across physics, chemistry, and biology.
\end{itemize}

\paragraph{Baseline Methods}
We compare our method against the following representative approaches designed for efficient reasoning:

\begin{itemize}
    \item \textbf{Concise Thoughts (CCoT)} \cite{renze2024ccot}: It encourages the model to generate concise reasoning process via simply append ``Be concise'' to the prompt. 
    
    \item \textbf{Chain of Draft (CoD)} \cite{xu2025chainofdraft}: This is another prompt-based method which instructs the model to generate concise draft intermediate steps during reasoning. \footnote{Please refer to the Appendix \ref{sec:appendix.2} for specific prompt templates for CCoT and CoD.}
  
    \item \textbf{SFT\textsubscript{Shortest}} \cite{munkhbat2025self,chen20242add3}: This method selects the shortest correct response from the backbone model's  sampled answers as the ground truth, and then performs supervised fine-tuning (SFT) on the backbone model.
    
    \item \textbf{SimPO\textsubscript{Shortest}} \cite{chen20242add3}: SimPO with contrastive instance pairs generated by the backbone reasoning model, which takes the shortest correct sampled response of each problem as positive instance and the longest correct counterpart as negative instance.

    \item \textbf{SimPO\textsubscript{Cosine}}:  We keep the DAST settings entirely unchanged, only replacing the ranking criterion for contrastive pairs from the reward function defined in Section \ref{subsection:reward} to the cosine reward function introduced in \cite{yeo2025demystifying}. We aim to verify the effectiveness of our proposed reward function through comparative analysis with \textbf{SimPO\textsubscript{Cosine}}.
    
   
    \item \textbf{SimPO\textsubscript{LenPenalty}}: We employ the length penalty reward function defined in \cite{team2025kimi} to evaluate the sampled responses for each question, select the highest and lowest ranked instances to construct contrastive pairs, and thereby develop another version of SimPO baseline.
    
\end{itemize}


\paragraph{Evaluation Metrics} We adopt following metrics to comprehensively assess both reasoning accuracy and compression effectiveness:
 \textbf{ACC} denotes the accuracy of the final answer. \textbf{LEN} refers to the average response length, measured in tokens. \textbf{C-LEN} represents the average number of tokens in all correct responses. \textbf{CR} is the compression ratio, which is defined as token length reduction ratio (vs. original model).  \textbf{C-CR} is the \textbf{C-LEN} reduction ratio against original model.


\paragraph{Training Details}

For both backbone reasoning models, we generate 20 candidate responses for each question in the MATH \cite{hendrycks2measuring} training set with maximum sequence length constrained to 4,096 tokens to compute its TLB. Following reward score calibration via Equation 2, we construct the preference training set for SimPO optimization. The truncation threshold $\delta$ is set to 0.15 and 0.18 for DS-7B and DS-32B, yielding final training sets of 10295 and 9813 contrastive pairs for DS-7B and DS-32B, respectively. All models are trained for 1 epoch using AdamW optimizer with learning rate lr = 5e-6. All our experiments were run on a NVIDIA GPU machine with 8 $\times$ H100. Please refer to Appendix \ref{sec:appendix.1} for complete training configuration statistics.


\paragraph{Decoding Configuration} In our evaluation setup, all models were  constrained to a maximum generation length of 32,768 tokens to align with DeepSeek' technical report \cite{guo2025deepseekr1}. Following \cite{chen20242add3,yeo2025demystifying}, we employ greedy decoding for all the models. Results were computed using  OpenR1 evaluation scripts\footnote{https://github.com/huggingface/open-r1}.


\begin{table*}[ht]
\centering
\renewcommand{\arraystretch}{1.25}
\resizebox{\textwidth}{!}{
\begin{tabular}{llcccccccccccccc ccc}
\toprule
\multirow{2}{*}{\textbf{MODEL}} & \multirow{2}{*}{\textbf{METHOD}} &
\multicolumn{5}{c}{\textbf{MATH-500}} &
\multicolumn{5}{c}{\textbf{AIME 2024}} &
\multicolumn{5}{c}{\textbf{GPQA}} &
 \\
 
\cmidrule(r){3-7} \cmidrule(r){8-12} \cmidrule(r){13-17} 

& &  ACC $\uparrow$  & LEN $\downarrow$ & C-LEN$\downarrow$ &  CR$\uparrow$ &  C-CR$\uparrow$  & ACC$\uparrow$ & LEN$\downarrow$ &  C-LEN$\downarrow$ & CR$\uparrow$ &  C-CR$\uparrow$  & ACC$\uparrow$ & LEN$\downarrow$ &  C-LEN$\downarrow$ & CR $\uparrow$ &  C-CR$\uparrow$  \\

\midrule
\multirow{8}{*}{DS-7B}
& Origin            & 93.2 & 4039& 3506 & -  & - & 60.0 & 10603 & 7448 & - & - & 47.98 & 8021 & 7242 & - & - \\
& CCoT     & 92.2 & 3388 & 2887 &  16.1\%  & 17.7\% & 40.0 & 10976 & 6497 & -3.5\% &  12.8\% & 47.98 & 7612 & 6914 & 5.1\% &  4.5\% \\
& CoD      & 75.8 & 1596 & 1234 & \textbf{60.5\%} & \textbf{64.8\%} & 43.3 & 9399 & 6630 & 11.4\% & 11.0\% & 49.49 & 7178 & 6932 & 10.5\% & 4.3\% \\
& SimPO\textsubscript{shortest}    & 89.8 & 1891 & 1557 & 53.2\% 
 & 55.6\% & 53.3 & 8291 & 3847 & 21.8\% & \textbf{48.4\%} & 50.51 & 6068 & 5401 & 24.4\%  & 25.4\%\\
& SFT\textsubscript{shortest}  & 91.8 & 2987 & 2408 & 26.0\% & 31.3\% & 50.0 & 12989 & 6227 & -22.5\% & 16.4\% & 51.01 & 7760 & 6578 & 3.3\% & 9.2\%\\
& SimPO\textsubscript{cosine}     & 93.2 & 3897 & 3223 & 3.5\% & 8.1\% & 63.3 & 12572 & 6783 & -18.6\%  & 8.9\% & 50.00 & 8230 & 6912 & -2.6\% & 4.6\%\\
& SimPO\textsubscript{LenPenalty}      & 89.4 & 1922 & 1612 & 52.4\% & 54.0\% & 63.3 & 7419 & 4478 & \textbf{30.0\%} & 39.9\% & 51.01 & 5860 & 4847 & \textbf{26.9\%} & \textbf{33.1\%}\\
& \textbf{DAST(ours)}   & \textbf{93.6} & 3309 & 2709 & 18.1\% & 22.8\% & \textbf{70.0} & 10804 & 7924 & -1.9\%  & \- 6.4\% &  \textbf{51.51} & 7684 & 6480 & 4.2\% & 10.5\% \\

\midrule

\multirow{8}{*}{DS-32B}
& Origin          & 94.4 & 3782 & 3384 & -  & -& 73.3 & 10955 & 9124 & -  & -& 65.15 & 6410 & 5923 & - & -\\
& CCoT    & 93.2 & 2044 & 1733 & 46.0\% & 48.8\% & 56.7 & 8436 & 5678 & 23.0\% & 37.8\% & 63.13 & 5820 & 5143 & 9.2\%  & 13.2\%	\\
& CoD    & 93.6 & 1941 & 1628 & 48.7\%  & 51.9\% & 43.3 & 7288 & 5065 & 33.5\%  & 44.5\% & 62.12 & 5107 & 4831 & 20.3\%  & 18.4\%\\
& SimPO\textsubscript{shortest}    & 89.0 & 1107 & 998 & \textbf{70.7\%}  & \textbf{70.5\%} & 36.7 & 2580 & 855 & \textbf{76.4\%}  & \textbf{90.6\%} & 63.13 & 2455 & 2286 & 61.7\%  & 61.4\%\\
& SFT\textsubscript{shortest}    & 94.6 & 2402 & 2141 & 36.5\%  & 36.7\% & 66.7 & 8204 & 6577 & 25.1\%  & 27.9\% & 64.65 & 6044 & 5030 & 5.7\%  & 15.1\%\\
& SimPO\textsubscript{cosine}      & 94.2 & 2325 & 1968 & 38.5\%  &\ 41.8\% & 63.3 & 7379 & 5317 & 32.6\%  & 41.7\% & \textbf{65.15} & 5835 & 4987 & 9.0\%  & 15.8\%\\
& SimPO\textsubscript{LenPenalty}      & 90.6 & 1190 & 1066 & 68.5\%  & 68.5\% & 43.3 & 2748 & 2047 & 74.9\%  & 77.6\% & 62.12 & 2375 & 2111 & \textbf{62.9\%}  & \textbf{64.4\%}	\\
    & \textbf{DAST(ours)}   &  \textbf{95.8} & 2044 & 1744 & 46.0\%  & 48.5\% & \textbf{76.7} & 7023 & 5409 & 35.9\%  & 40.7\% & \textbf{65.15} & 5535 & 4514 & 13.7\%  & 23.8\%\\

\bottomrule
\end{tabular}}
\caption{Evaluation results across the benchmarks.}
\label{tab:all_eval}
\end{table*}

\subsection{Results and Analysis}

\subsubsection{Overall Results}

The main results are presented in Table \ref{tab:all_eval}. We have the following findings:
Prompt-based methods (CCoT, CoD) show unstable performance, often incurring accuracy losses, which are particularly pronounced on complex task AIME 2024. For example, ACC of CoD with DS-7B on AIME 2024 drops from 60.0\% to 43.3\%. Aggressive compression methods (SimPO\textsubscript{Shortest}, SimPO\textsubscript{LenPenalty}) achieve the most significant token reduction on both DS-7B and DS-32B across all benchmarks. However, this substantial compression invariably sacrifices some accuracy. SimPO\textsubscript{LenPenalty} demonstrates a slightly better overall balance against SimPO\textsubscript{Shortest} , potentially because its reward function introduces the average length of batch data as a comparison baseline, thereby better navigating the length-accuracy trade-off. SFT\textsubscript{Shortest} proves to be a strong baseline, especially with the DS-32B model. However, notably on the AIME 2024, SFT\textsubscript{Shortest} fails to reduce output length effectively, it is plausible that the straightforward SFT with shortest responses may have compromised the model's instruction following ability, resulting in ineffective termination of responses when confronted with complex tasks such as AIME 2024.

\textsc{DAST} and SimPO\textsubscript{Cosine} exhibit similar overall trends in balancing ACC and CR, the potential reason may be that neither method strictly prioritizes brevity but can encourage longer responses when beneficial. The superior performance of \textsc{DAST} over the standard cosine-based reward across benchmarks on both ACC and CR validates the effectiveness of our proposed budget-based reward function.

Despite being exclusively math-trained, DAST (DS-7B) achieves 51.51\% (+3.53\%) on GPQA with modest compression ratio (4.2\%), demonstrating certain ability of domain generalization. On the challenging AIME 2024 benchmark, DAST (DS-7B) does not reduce the average response length (CR -1.9\%). This, combined with a substantial accuracy improvement from 60.0\% (Origin) to 70.0\% suggests that \textsc{DAST} does not indiscriminately shorten reasoning paths but can adaptively allocate more reasoning steps for complex problems.

Overall, the results in Table \ref{tab:all_eval} affirm that \textsc{DAST} effectively navigates the intricate trade-off between conciseness and reasoning performance. It generally preserves or improves the reasoning capabilities of the backbone models while achieving remarkable CoT reductions, outperforming the baselines in this combined objective. This is particularly evident with the more capable DS-32B model, where DAST achieves strong compression alongside ACC improvements across all the benchmarks.

\begin{figure*}[htbp]
\centering
\begin{subfigure}[b]{0.49\textwidth} 
    \includegraphics[width=\textwidth]{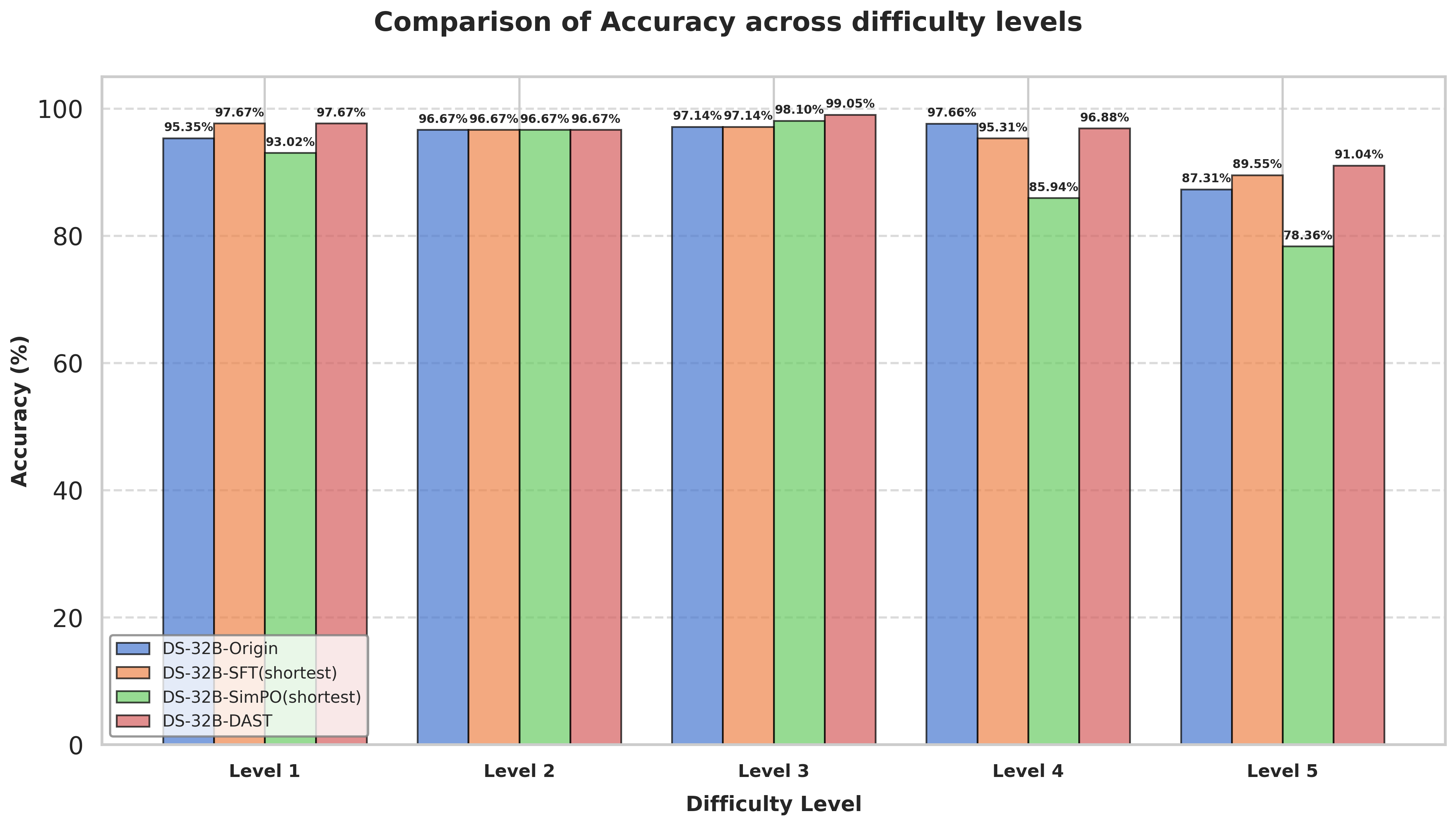}
    \caption{ Accuracy on different levels. }
    \label{fig:image-a}
\end{subfigure}
\hfill 
\begin{subfigure}[b]{0.49\textwidth} 
    \includegraphics[width=\textwidth]{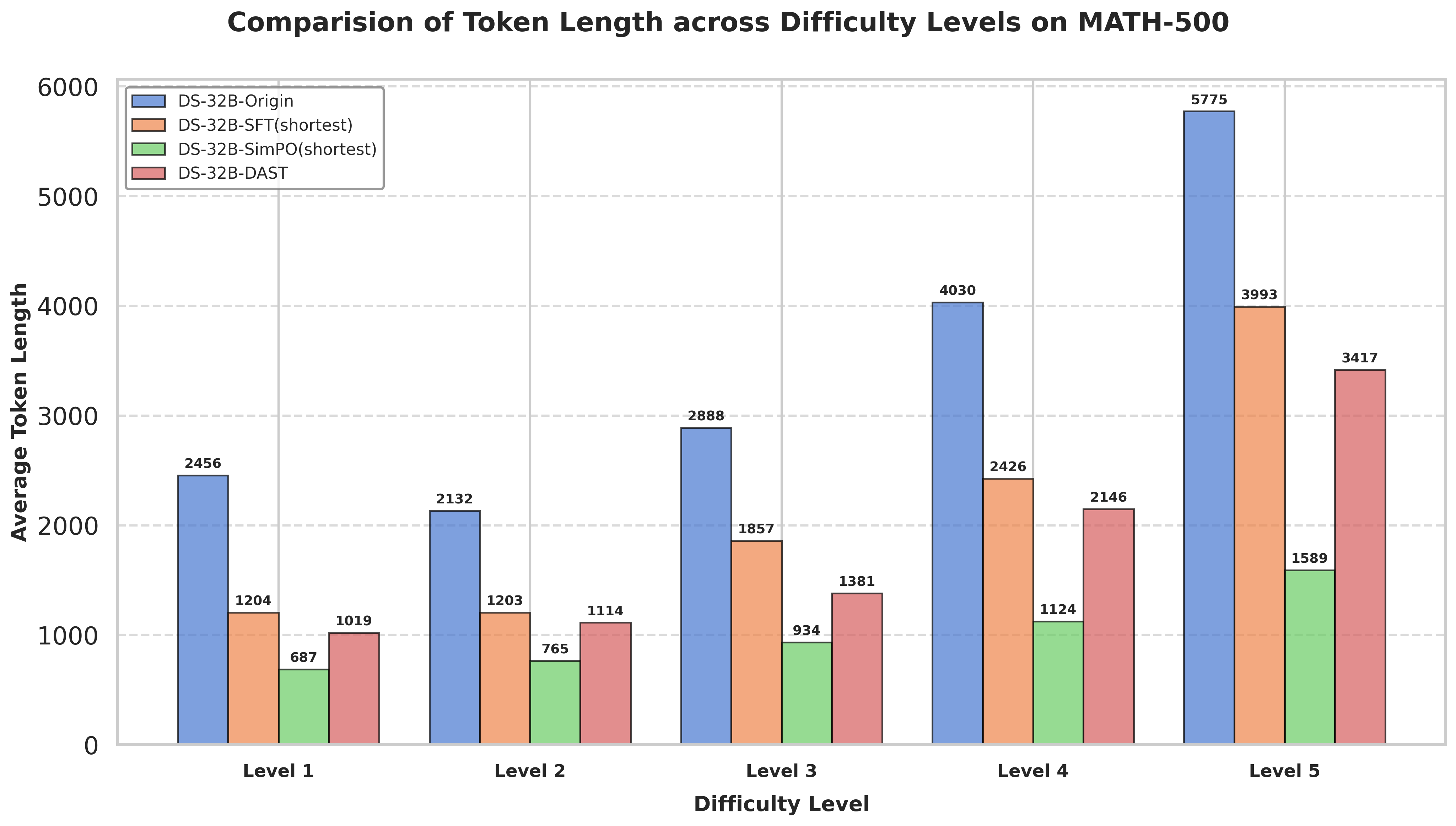}
    \caption{Token length on different levels .}
    \label{fig:image-b}
\end{subfigure}
\caption{Comparative results for different difficulty levels on MATH-500}
\label{fig:two-images}
\end{figure*}

\subsubsection{Fine-grained Analysis}

We also compared CR of DAST (DS-32B) on MATH-500 according to difficulty level. As shown in Figures \ref{fig:two-images}, DAST achieved the best Level~5 accuracy with significant margin against other methods, which demonstrates that it maintains its reasoning ability under complex problems. Although SimPO\textsubscript{Shortest} shows the most significant reduction in response length, its reasoning capability notably declines when addressing complex problems.

\begin{table}[ht]
\centering
\scalebox{0.82}{
\begin{tabular}{lccccc}
\toprule
\textbf{Model} & \textbf{L1} &  \textbf{L2} &  \textbf{L3} & \textbf{L4} & \textbf{L5} \\
\midrule
SimPO\textsubscript{Shortest} & 72.0\% & 64.1\% & 67.6\% & 72.1\% & 72.5\%\\
DAST & 58.5\% & 47.7\% & 51.9\% &  46.7\% & 40.8\%\\

\bottomrule
\end{tabular}}
\caption{Comparison of compression ratio 
between SimPO\textsubscript{Shortest} and DAST across different  levels in MATH-500 with DS-32B.}
\label{tab:relative_reduction}
\end{table}

Statistical analysis of compression ration at different levels on MATH-500 in Table \ref{tab:relative_reduction} reveals that SimPO\textsubscript{Shortest} exhibits limited differentiation in CR across different difficulty levels. In contrast, the DAST method shows discernible adaptive capabilities, achieving approximately 58.5\% length reduction at Level~1 compared to the original model, while this reduction decreases to approximately 40.8\% at the most challenging Level~5. This progressive performance variation validating its difficulty-adaptive nature.

\subsubsection{Ablation Study}
\label{subsection:ablation}

To reveal the individual effects of different components of our method, we tested different variants of DAST on MATH-500 with DS-7B by removing DCP or DICP. The ablation results are shown in Table \ref{tab:ablation}. 
We see that the DCP and DICP components exhibit specialization patterns analogous to domain-specific experts. When removing DCP while retaining DICP (w/o DCP), the framework incentivizes models to fully utilize the token budget, resulting in an accuracy improvement(+1.4\% versus DS-7B). However, this comes at the cost of overly redundant answer length (+17.8\% versus DS-7B). Conversely, eliminating DICP while preserving DCP (w/o DICP) drives the model to strictly adhere to budget constraints through aggressive compression, achieving optimal compression ratio (59.8\%) but significantly impairing problem-solving capability (-3.2\% accuracy). The optimal performance is achieved when DCP and DICP are combined, indicating that DCP and DICP are complementary to each other. We further explored integrating CICP into DAST's training set (incorporating CICPs with the largest reward score discrepancies per question). However, as evidenced in Table \ref{tab:ablation} (bottom row), this integration yielded no significant performance gains. We will investigate the ineffectiveness of CICP in the future work.

\begin{table}[ht]
\centering
\scalebox{0.82}{
\begin{tabular}{lcccc}
\toprule
\textbf{Model} & \textbf{ACC} &  \textbf{LEN} &  \textbf{C-LEN} & \textbf{CR} \\
\midrule
DS-7B & 93.2 & 4039.13 & 3506.64 & -\\
\midrule
DAST & 93.6 & 3309.16 & 2708.63 &  18.0\%\\
w/o DCP & 94.6 & 4759.07 & 3996.78 &  -17.8\% \\
w/o DICP & 90.0 & 1624.60 & 1299.50 & 59.8\% \\
+ CICP & 93.2 & 3295.96 & 2573.00 & 18.3\% \\
\bottomrule

\end{tabular}}
\caption{Ablation Results on MATH-500 with DS-7B.}
\label{tab:ablation}
\end{table}

\subsubsection{Impact of Truncation Threshold }

\begin{figure}[htbp]
\centering
\includegraphics[width=0.48\textwidth]{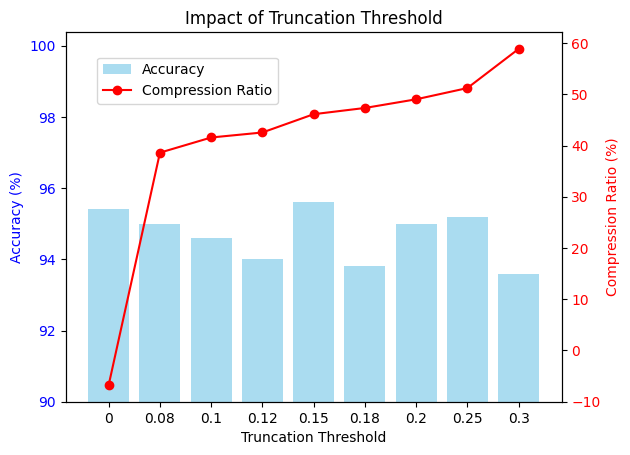}
\caption{The impact of truncation threshold $\delta$.} \label{fig:threshold}
\end{figure}

To investigate the impact of the truncation threshold $\delta$, we conducted grid search validation on 100 randomly selected samples from MATH-500. As shown in Figure \ref{fig:threshold}, the DS-32B model achieves peak ACC with  $\delta$ = 0.15, accompanied by a CR of 47\% in generated token length. This empirical evidence guided our final selection of $\delta$ = 0.15 for DS-32B to optimize the Accuracy. For the DS-7B variant, the same hyperparameter search on the same validation set identified 0.18 as the optimal $\delta$.

It is noteworthy that when  $\delta$ = 0, the model's CR becomes extremely low (even negative), primarily because the training data contains DICPs with low discriminability and excessive length, which causes reward hacking and prevents SimPO from capturing the correct direction for length optimization.

\section{Conclusion}

This work addresses the critical efficiency-performance dilemma in slow thinking models through difficulty-aware reasoning adaptation. By establishing correlation between problem complexity and optimal solution length, the proposed DAST framework enables dynamic resource allocation for reasoning. Experimental validations across representative benchmarks confirm the effectiveness of our method.

\section*{Limitations}

While our introduced method achieves a remarkable trade-off between reasoning accuracy and response compression rate, following limitations warrant discussion:

\paragraph{Domain-Specific Evaluation Scope} Our current benchmarking focuses exclusively on STEM disciplines (e.g., mathematics, physics, chemistry), leaving code generation and general domain tasks unexplored. We plan to extend the evaluation benchmarks in the future.

\paragraph{Threshold Sensitivity} Our method is sensitive to the truncation threshold. Therefore, it requires some additional cost to carefully adjust the threshold.

\paragraph{Off-Policy Learning Constraints} The proposed DAST framework, though computationally efficient through preconstructed training data, may inherently limit performance potential compared to online reinforcement learning approaches. 
We plan to explore on-policy reinforcement learning variants using our designed reward function for further improvement.

\bibliography{acl_latex}

\newpage

\appendix

\section{Implementation Details}
\label{sec:appendix.1}

\begin{table}[ht]
\centering
\scalebox{0.82}{
\begin{tabular}{lll}
\toprule
\textbf{Model} & \textbf{Name} & \textbf{Value} \\
\midrule
\multirow{8}{*}{DS-7B} & training samples & 10295 \\
                        & learning rate & 5e-6  \\
                        & DCP \%          & 91.26\%   \\
                        & DICP \%         & 8.74\%   \\
                        & epoch          & 1      \\
                        & $L_{max}$      & 4096   \\
                        & $\beta$        & 200    \\
                        & $\gamma$       & 1      \\

\midrule
\multirow{8}{*}{DS-32B} & training samples & 9813  \\
                        & learning rate & 5e-6   \\
                        & DCP \%          & 87.73\%   \\
                        & DICP \%         & 12.27\%   \\
                        & epoch          & 1      \\
                        & $L_{max}$      & 4096   \\
                        & $\beta$        & 200    \\
                        & $\gamma$       & 1      \\

\bottomrule

\end{tabular}}
\caption{Training configuration of DAST.}
\label{tab:parameters}
\end{table}

\section{Prompt Templates of CCoT and CoD}

\begin{figure}[htbp]
\centering
\includegraphics[width=0.48\textwidth]{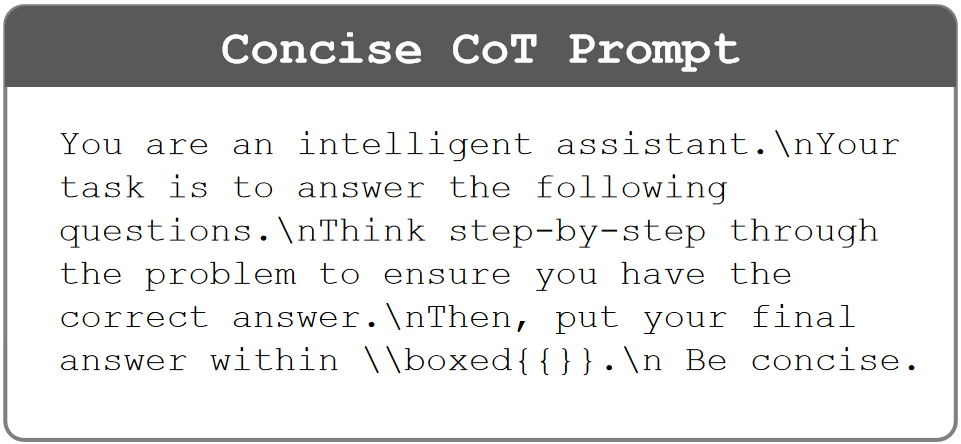}
\caption{The prompt we used to implement CCoT method.} \label{fig:ccot}
\end{figure}

\begin{figure}[htbp]
\centering
\includegraphics[width=0.48\textwidth]{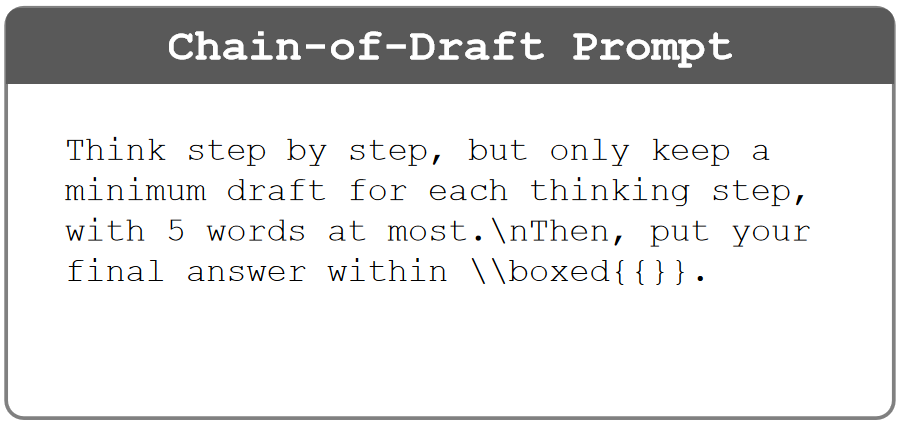}
\caption{The prompt we used to implement CoD method.} 
\label{fig:cod}
\end{figure}

\label{sec:appendix.2}

\section{Case Study}
\label{sec:appendix.3}

\begin{figure}[htbp]
\centering
\includegraphics[width=0.5\textwidth]{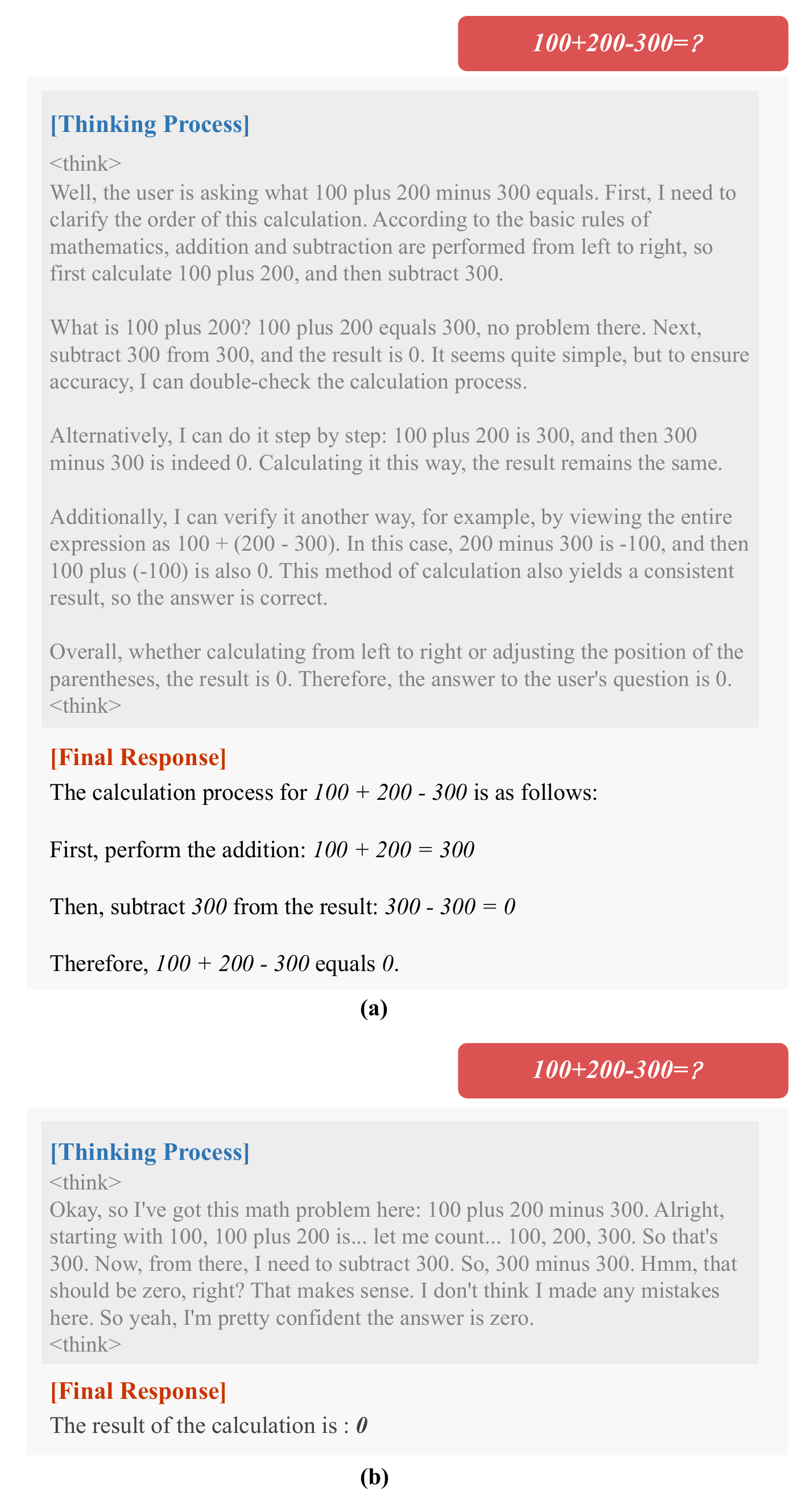}
\caption{(a) An overthinking case from DeepSeek-R1-Distill-Qwen-32B. (b) Output from DeepSeek-R1-Distill-Qwen-32B with DAST.} 
\label{fig:case1-2}
\end{figure}

Figure \ref{fig:case1-2} demonstrate a comparison of results for a simple problem from DeepSeek-R1-Distill-Qwen-32B. It can be observed that that the original outputs include extensive unnecessary and redundant thinking processes, while the outputs after applying DAST are more concise and focused.


\end{document}